\documentclass{article}
\usepackage{aaai}
\usepackage{epsfig}
\usepackage{xspace}
\usepackage{url}

\newcommand{\nop}[1]{}

\newcommand{\dlv}{{\bf\small{DLV}}\xspace}
\newcommand{\gc}{\ensuremath{\textbf{Guess\&Check}}}

\newcommand{\Or}{\ensuremath{\mathtt{v}}}
\newcommand{\derives}{\ensuremath{\mathtt{\mbox{\texttt{:-}}}}}

\newcommand{\tneg}{\ensuremath{-}}
\newcommand{\naf}{\ensuremath{\mathtt{not}}}

\newcommand{\p}{\ensuremath{{\cal P}}}
\newcommand{\R}{\ensuremath{r}}
\newcommand{\HR}{\ensuremath{H(\R)}}
\newcommand{\BR}{\ensuremath{B(\R)}}
\newcommand{\BpR}{\ensuremath{B^+(\R)}}
\newcommand{\BnR}{\ensuremath{B^-(\R)}}

\newcommand{\GP}{\ensuremath{Ground(\p)}}
\newcommand{\GR}{\ensuremath{Ground(\R)}}
\newcommand{\BP}{\ensuremath{B_{\p}}}
\newcommand{\UP}{\ensuremath{U_{\p}}}

\newcommand{\NP}{{\rm NP}}
\newcommand{\SigmaP}[1]{{\Sigma}_{#1}^{P}}

\begin{document}

\title{\mbox{DLV}\ -- A System for 
Declarative \\
Problem Solving\thanks{%
This work was supported by FWF (Austrian Science Funds) under the
projects \mbox{P11580-MAT} and \mbox{Z29-INF}.
}}

\author{\mbox{Thomas Eiter}\and %
\mbox{Wolfgang Faber} \and %
\mbox{Christoph Koch} \AND %
\mbox{Nicola Leone}\and %
\mbox{Gerald Pfeifer}
\\[0.9ex]
Institut f{\"u}r Informationssysteme, TU Wien\\
Favoritenstrasse 9-11\\ A-1040 Wien, Austria\\
\{leone,pfeifer\}@dbai.tuwien.ac.at, \{eiter,faber\}@kr.tuwien.ac.at 
\\[0.9ex]
European Organization for Nuclear Research\\
CERN EP Division\\
CH-1211 Geneva, Switzerland\\
christoph.koch@cern.ch
}

\maketitle

\begin{abstract}
\noindent 
\dlv is an efficient logic programming and non-monotonic reasoning (LPNMR)
system with advanced knowledge representation mechanisms and
interfaces to classic relational database systems.

Its core language is disjunctive datalog (function-free disjunctive
logic programming) under the Answer Set Semantics with integrity
constraints, both default and strong (or explicit) negation, and
queries. Integer arithmetics and various built-in predicates are also
supported.

In addition \dlv has several frontends, namely brave and cautious
reasoning, abductive diagnosis, consistency-based diagnosis, a subset
of SQL3, planning with action languages, and logic programming with
inheritance.
\end{abstract}

\section{General Information}


Currently \dlv is available in binary form for various platforms
(sparc-sun-solaris2.6, alpha-dec-osf4.0, i386-linux-elf-gnulibc2,
i386-pc-solaris2.7, and i386-unknown-freebsdelf3.3 as of this writing)
and it is easy to build \dlv on further platforms.

Including all frontends, \dlv consists of around 25000 lines ISO C{\small ++}
code plus several scanners and parsers written in lex/flex and
yacc/bison, respectively. \dlv is being developed using GNU tools
(GCC, flex, and bison) and is therefore portable to most Unix-like
platforms. Additionally, the system has been successfully built with
proprietary compilers such as those of Compaq and SCO.

For up-to-date information on the system and a full manual please
refer to the project homepage \cite{dlv-web}, where you can also
download \dlv.

\section{Description of the System} 

\subsection{Kernel Language}

The kernel language of \dlv is disjunctive datalog extended with strong
negation under the answer set semantics \cite{eite-etal-97f,gelf-lifs-91}. 

\subsubsection*{Syntax}

Strings starting with uppercase letters denote variables, while those
starting with lower case letters denote constants. A {\em term} is
either a variable or a constant.  An {\em atom} is an expression
$p(t_{1},\ldots$,$t_{n})$, where $p$ is a {\em predicate} of arity $n$
and $t_{1}$,\ldots,$t_{n}$ are terms.  A {\em literal} $l$
is either an atom $a$ (in this case, it is {\em positive}), or a
negated atom $\tneg a$ (in this case, it is {\em negative}).

Given a literal $l$, its {\em complementary} literal is
defined as $\tneg a$ if $l=a$ and $a$ if $l=\tneg a$.  A set $L$
of literals is said to be {\em consistent} if for every
literal $l \in L$, its complementary literal is not contained in $L$.

In addition to literals as defined above, \dlv also supports
built-ins, like \texttt{\#int}, \texttt{\#succ}, \texttt{$<$}, \texttt{+},
and \texttt{*}. For details, we refer to our full manual \cite{dlv-web}.

A {\em disjunctive rule} ({\em rule}, for short) \R{} is a formula
\[
a_1\ \Or\ \cdots\ \Or\ a_n\ \derives\ 
        b_1,\cdots, b_k,\ 
        \naf\ b_{k+1},\cdots,\ \naf\ b_m.
\]
where $a_1,\cdots ,a_n,b_1,\cdots ,b_m$ are literals,
$n\geq 0,$ $m\geq k\geq 0$, and $\naf$ represents \emph{negation-as-failure}
(or \emph{default negation}).
The disjunction $a_1\Or\cdots\Or a_n$
is the {\em head} of \R{}, while the conjunction $b_1 , ..., b_k,\
\naf\ b_{k+1} , ...,\ \naf\ b_m$ is the {\em body} of \R{}. A rule
without head literals (i.e.\ $n=0$) is usually referred to as {\em
integrity constraint}. If the body is empty (i.e.\ $k=m=0$), we
usually omit the ``\derives{}'' sign.

We denote by $H(r) \nop{=\{a_1,\ldots$, $a_n \}}$ the set of
literals in the head, and by $\BR{}=\BpR \cup \BnR$ the set of the
body literals, where $\BpR=$ $\{b_1,$\ldots, $b_k\}$ and
$\BnR=\{b_{k+1},$ \ldots, $b_m \}$ are the sets of positive and
negative body literals, respectively.

A {\em disjunctive datalog program} \p{} is a finite set of rules.


\nop{
As usual, an object (term, atom, etc) is called {\em ground}, if no
variables appear in it. A ground program is also called a {\em
propositional} program.

Sometimes, we speak about {\em general} programs, if we want to point
out that they are not restricted to be positive, normal or ground.
}

\subsubsection*{Semantics}

\dlv implements the consistent answer sets semantics which has originally
been defined in \cite{gelf-lifs-91}.\footnote{Note that we only
consider {\em consistent answer sets}, while in \cite{lifs-96} also
the inconsistent set of all possible literals is a valid answer set.}

Before we are going to define this semantics, we need a few
prerequisites. As usual, given a program \p{}, \UP{} (the
\emph{Herbrand Universe}) is the set of all constants appearing
in \p{} and \BP{} (the \emph{Herbrand Base}) is the set of all
possible combinations of predicate symbols appearing in \p{} with
constants of \UP{} possibly preceded by \tneg, in other words, the set
of ground literals constructible from the symbols in \p.

Given a rule \R{}, \GR{} denotes the set of rules obtained by applying
all possible substitutions $\sigma$ from the variables in \R{} to
elements of \UP{}; \GR{} is also called the \emph{Ground
Instantiation} of \R{}. In a similar way, given a program \p{}, \GP{}
denotes the set \( \displaystyle
\bigcup_{\R \in \p} \GR \). For programs not containing variables \( \p{} = \GP{}
\) holds.

For every program \p{}, we define its \emph{answer sets} using its
ground instantiation \GP{} in two steps, following \cite{lifs-96}:
First we define the answer sets of positive programs, then we give a
reduction of general programs to positive ones and use this reduction
to define answer sets of general programs.

An interpretation $I$ is a set of literals.  A consistent
interpretation $I \subseteq \BP$ is called {\em closed under}
a positive, i.e.\ \naf-free, program \p, if, for every $\R
\in \GP$, $\HR \cap I \neq \emptyset$ whenever $\BR \subseteq I$. $I$
is an {\em answer set} for a positive program \p{} if it is minimal
w.r.t.\ set inclusion and closed under \p{}.

The {\em reduct} or {\em Gelfond-Lifschitz transform} of a general
ground program \p{} w.r.t.\ a set $X \subseteq \BP$ is the positive
ground program $\p^X$, obtained from \p{} by deleting all rules $\R
\in \p$ for which $\BnR \cap X \neq \emptyset$ holds, and deleting the
negative body from the remaining rules.

An answer set of a general program \p{} is a set $X \subseteq \BP$ such
that $X$ is an answer set of $\GP^X$.

\subsection{Application Frontends}

In addition to its kernel language, \dlv provides a number of application
frontends that show the suitability of our formalism for solving various
problems from the areas of Artificial Intelligence, Knowledge Representation
and (Deductive) Databases.

\begin{itemize}

\item The \emph{Brave and Cautious Frontends} are simple extensions of the
normal mode, where in addition to a disjunctive datalog program the user specifies a
conjunction of literals (a query) and \dlv checks whether this query holds
in any respectively all answer sets of the program.

\item The \emph{Diagnoses Frontend} implements both abductive
diagnosis \cite{pool-89,cons-etal-91}, adapted to the semantics of
logic programming \cite{kaka-etal-93,eite-etal-97k}, and
consistency-based diagnosis \cite{reit-87,dekl-etal-92} and supports
general diagnosis as well as single-failure and subset-minimal
diagnosis.

\item The \emph{SQL3 Frontend} is a prototype implementation of the
query interface of the SQL3 standard that has been approved by ISO
last year.

\item The \emph{Inheritance Frontend} extends the kernel language of
\dlv with objects, and inheritance \cite{bucc-etal-99a-iclp}.
This extension allows us to naturally represent inheritance and
default reasoning with (multi-level) exceptions, providing a natural
solution also to the frame problem.

\item Finally, the \emph{Planning Frontend} implements a
new logic-based planning language, called \ensuremath{\cal K}, which
is well suited for planning under incomplete knowledge.

\end{itemize}

\subsection{Architecture}

\begin{figure}
\centering
\epsfig{file=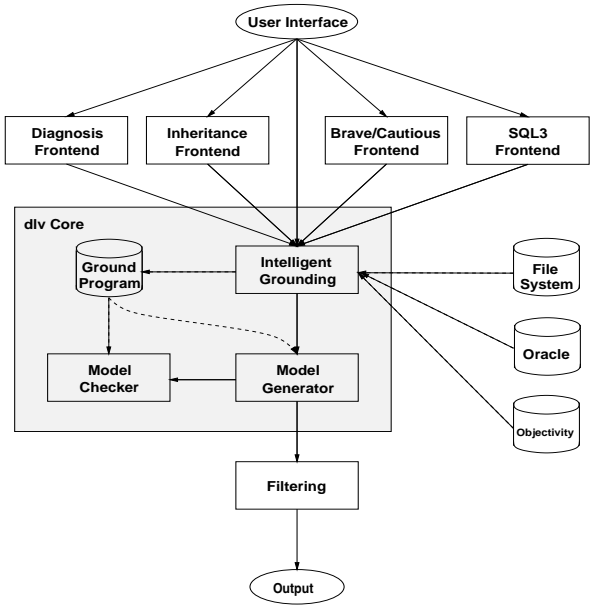,height=8cm,width=7.8cm}
\caption{Overall architecture of \dlv.}
\label{f-architecture}
\end{figure}

An outline of the general architecture of our system is depicted in
Fig.\ref{f-architecture}.

The heart of the system is the {\em \dlv core}. Wrapped around this
basic block are frontend preprocessors and output filters (which also
do some post-processing for frontends). The system takes input data
from the user (mostly via the command line) and from the file system
and/or database systems.

Upon startup, input is possibly translated by a frontend.  Together
with relational database tables, provided by an Oracle database, an
Objectivity database, or ASCII text files, the {\em Intelligent
Grounding Module},
efficiently generates a subset of the grounded input program that has
exactly the answer sets as the full program, but is much smaller in
general.

After that, the Model Generator is started. It generates one answer set
candidate at a time and verifies it using the Model Checker. Upon
success, filtered output is generated for the answer set. This process
is iterated until either no more answer sets exist or an explicitly
specified number of answer sets has been computed.

Not shown in Fig.\ref{f-architecture} are various additional
data structures, such as dependency graphs.

\section{Applying the System}


\subsection{Methodology}

                                        

The core language of \dlv can be used to encode problems in a highly
declarative fashion, following a ``$\gc$'' paradigm.  We will first
describe this paradigm in an abstract way and then provide some
concrete examples.  We will see that several problems, also problems
of high computational complexity, can be solved naturally in \dlv by
using this declarative programming technique. The power of disjunctive
rules allows one to express problems, which are even more complex than
\NP\ uniformly over varying instances of the problem using a fixed
program.

Given a set $F_I$ of facts that specify an instance $I$ of some
problem $P$, a $\gc$ program $\p{}$ for $P$ consists of the following two
parts:
\begin{description}
\item[Guessing Part] The guessing part $G \subseteq \p{}$ 
defines the search space, in a way such that answer sets of $G\cup
F_I$ represent ``solution candidates'' of $I$.
\item[Checking Part] The checking part $C \subseteq \p{}$ tests
whether a solution candidate is in fact a solution, such that the
answer sets of $G\cup C \cup F_I$ represent the solutions for the
problem instance $I$.
\end{description}

In general, we may allow both $G$ and $C$ to be arbitrary collections of
rules in the program, and it may depend on the complexity of
the problem which kind of rules are needed to realize these parts (in
particular, the checking part); we defer this discussion to a later
point in this section. 

Without imposing restrictions on which rules $G$ and $C$ may contain,
in the extremal case we might set $G$ to the full program and let $C$
be empty, i.e., all checking is moved to the guessing part such that
solution candidates are always solutions. This is certainly not
intended. However, in general the generation of the search space may
be guarded by some rules, and such rules might be considered more
appropriately placed in the guessing part than in the checking
part. We do not pursue this issue any further here, and thus also
refrain from giving a formal definition of how to separate a program
into a guessing and a checking part.

For solving a number of problems, however, it is possible to design a
natural $\gc$ program in which the two parts are clearly identifiable
and have a simple structure:
\begin{itemize}
\item
The guessing part $G$ consists of a disjunctive rule which ``guesses'' a
solution candidate $S$.
\item
The checking part $C$ consists of integrity constraints which check
the admissibility of $S$, possibly using auxiliary predicates which
are defined by normal stratified rules.
\end{itemize}

In a sense, the disjunctive rule defines the search space in which rule
applications are branching points, while the integrity constraints
prune illegal branches.

As a first example, let us consider \emph{Hamiltonian Path}, a
classical $\NP$-complete problem from graph theory.

\paragraph{HPATH:}
Given a directed graph $G=(V,E)$ and a vertex $a$ of this graph,
does there exist a path of $G$ starting at $a$ and passing through
each vertex in $V$ exactly once?

Suppose that the graph $G$ is specified by means of predicates $node$
(unary) and $arc$ (binary), and the starting node is
specified by the predicate $start$ (unary).
Then, the following $\gc$ program $\p_{hp}$ solves the Hamilton Path
problem.

\medskip

{\small\tt 
$\begin{array}{l}\setlength{\arraycolsep}{0pt}
\left.
\begin{minipage}{\textwidth}
\begin{tabbing}
$\left.
\begin{array}{l}
\mbox{
\derives{} inPath(X,Y), inPath(X,Y1), Y $<>$ Y1.
}
\end{array}
\right\} \mathbf{Constraints}\!$\=\kill
inPath(X,Y) \Or{} outPath(X,Y) \derives{} arc(X,Y).\>
\end{tabbing}
\end{minipage}
\right\} \mathbf{Guess}
\\*[2ex]
\left.
\begin{array}{l}
\left.
\begin{minipage}{\textwidth}
\begin{tabbing}
\derives{} \= inPath(X,Y), inPath(X,Y1), Y $<>$ Y1.\\
\derives{} \> inPath(X,Y), inPath(X1,Y), X $<>$ X1.\\
\derives{} \> node(X), \naf\ reached(X).
\end{tabbing}
\end{minipage}
\right. \\*[1ex]
\left.
\begin{minipage}{\textwidth}
\begin{tabbing}
\= \derives{} inPath(X,Y), inPath(X,Y1), Y $<>$ Y1.\=\kill
reached(X) \= \derives{} start(X).\>\\
reached(X) \> \derives{} reached(Y), inPath(Y,X).
\end{tabbing}
\end{minipage}
\right. 
\end{array} \right\} C
\end{array}$
}

\medskip

The first rule guesses a subset of all given arcs, while the
rest of the program checks whether it is a Hamiltonian Path. Here, the
checking part $C$ uses an auxiliary predicate $\mathtt{reached}$,
which his defined using positive recursion. 

In particular, the first two constraints in $C$
check whether the set of arcs $S$ selected by $\mathtt{inPath}$ meets
the following requirements, which any Hamiltonian Path must satisfy:
There must not be two arcs
starting at the same node, and there must not be two arcs ending in
the same node.  

The two rules after the constraints define reachability from the
starting node with respect to the selected arc set $S$. This is used
in the third constraint, which enforces that all nodes in the graph
are reached from the starting node in the subgraph induced by
$S$. This constraint also ensures that this subgraph is connected.

It is easy to see that a selected arc set $S$ which satisfies all
three constraints must contain the edges of a path
$a=v_0,v_1,\ldots,v_k$ in $G$ that starts at node $a$, and passes
through distinct nodes until no further node is left, or it arrives at
the starting node $a$ again. In the latter case, this means that the
path is a Hamiltonian Cycle, and by dropping the last edge, we
have a Hamiltonian Path.

Thus, given a set of facts $F$ for $node$, $arc$, and $start$ which
specify the problem input, the program $\p_{hp}\cup F$ has an answer
set if and only if the input graph has a Hamiltonian Path.

If we want to compute a Hamiltonian Path rather than only answering
that such a path exists, we can strip off the last edge from a
Hamiltonian Cycle by adding a further constraint
\texttt{\derives{} start(Y), inPath(\_,Y).} to the program. Then, the
set $S$ of selected edges in an answer sets of $\p_{hp}\cup F$
constitutes a Hamiltonian Path starting at $a$.

It is worth noting that \dlv is able to solve problems which are
located at the second level of the polynomial hierarchy, and indeed
also such problems can be encoded by the $\gc$ technique, as in the
following example called \emph{Strategic Companies}.

\paragraph{STRATCOMP:}
Given the collection $C= \{c_1$, \ldots $c_m\}$ of companies $c_i$
owned by a holding, and information about company control, compute the
set of the strategic companies in the holding.

To briefly explain what ``strategic'' means in this context, imagine
that each company produces some goods.  Moreover, several companies
jointly may have control over another company. Now, some companies
should be sold, under the constraint that all goods can be still
produced, and that no company is sold which would still be controlled
by the holding after the transaction. A company is {\em strategic}, if
it belongs to a {\em strategic set}, which is a minimal set of
companies satisfying these constraints.

This problem is $\SigmaP{2}$-hard in general \cite{cado-etal-97};
reformulated as a decision problem (``Given a further company $c$ in the
input, is $c$ strategic?''), it is $\SigmaP{2}$-complete. To our
knowledge, it is the only KR problem from the business domain of this
complexity that has been considered so far.

In the following encoding, $\mathtt{strat(X)}$ means that $\mathtt X$
is strategic, $\mathtt{company(X)}$ that $\mathtt{X}$ is a company,
$\mathtt{produced\_by(X,Y,Z)}$ that product $\mathtt X$ is produced by
companies $\mathtt Y$ and $\mathtt Z$, and
$\mathtt{controlled\_by(W,X,Y,Z)}$ that $\mathtt W$ is jointly
controlled by $\mathtt{X,Y}$ and $\mathtt Z$. We have adopted the
setting from \cite{cado-etal-97} where each product is produced by at
most two companies and each company is jointly controlled by at most
three other companies.

Given the facts $F$ for $company$, $controlled\_by$ and
$produced\_by$, the answer sets of the following program $\p{1}$
(actually $\p{1}\cup F$) correspond one-to-one to the strategic sets
of the holding. Thus, the set of all strategic companies is given by
the set of all companies $\mathtt{c}$ for which the fact
$\mathtt{strat(c)}$ is true under brave reasoning.

\[
\begin{array}{l}
\left.
\begin{array}{l}
r:\ \mathtt{strat(Y)\ \Or{}\ strat(Z)\ \mbox{\texttt{\derives{}}}\ produced\_by(X,Y,Z).}
\end{array} \right.
\\[1ex]
\left.
\begin{array}{l@{~}cl@{~}}
c:\ \mathtt{strat(W)}\ & \texttt{\derives{}}& \mathtt{controlled\_by(W,X,Y,Z),}\\
 && \mathtt{strat(X),\ strat(Y),\ strat(Z).}
\end{array}
\right.
\end{array}
\]

\vspace{0.2cm}

Intuitively, the guessing part $G$ of $\p{1}$ consists of the
disjunctive rule $r$, and the checking part $C$ consist of the normal
rule $c$. This program exploits the minimization which is inherent to
the semantics of answer sets for the check whether a candidate set $S$
of companies that produces all goods and obeys company control
is also minimal with respect to this property.

The guessing rule $r$ intuitively selects one of the companies
$\mathtt{c_1}$ and $\mathtt{c_2}$ that produce some item $\mathtt{g}$,
which is described by $\mathtt{produced\_by(g,c_1,c_2)}$.  If there
were no company control information, minimality of answer sets would
then naturally ensure that the answer sets of $F \cup \{r \}$
correspond to the strategic sets; no further checking is needed.
However, in case such control information, given by facts
$\mathtt{controlled\_by(c,c_1,c_2,c_3)}$, is available, the rule $c$
in the program checks that no company is sold that would be controlled
by other companies in the strategic set, by simply requesting that
this company must be strategic as well. The minimality of the strategic
sets is automatically ensured by the minimality of answer sets. The
answer sets of $\p{2}$ correspond one-to-one to the strategic sets of
the given instance.

It is interesting to note that the checking constraint $c$ interferes
with the guessing rule $r$: applying $c$ may spoil the minimal answer
set generated by rule $r$. Such feedback from the checking part $C$ to
the guessing part $G$ is in fact needed to solve $\SigmaP{2}$-hard
problems.

\nop{STRATCOMP, if $G$ does not contain disjunctive
rules. This follows from complexity considerations: if $G$ had no
influence on $C$ (and $F$), the answer sets of $G \cup C \cup F$ would
just be the answers sets of $C \cup X$, where $X$ is an answer set of
$G \cup F$. The notion of ``influence'' can be made precise in terms
of a ``potentially uses'' relation \cite{eite-etal-94a,eite-etal-97f}
and a ``splitting set'' \cite{lifs-turn-94}. However, along with $X$
an answer set $Y$ of $C \cup X$ can be guessed and, if $C$ contains no
disjunctive rule, both $X$ and $Y$ can be verified in polynomial time. This
would mean complexity in $\NP$---this is impossible unless
$\SigmaP{2}$ = $\NP$, which is believed to be false by the experts.
}

In general, if a program encodes a problem that is
$\SigmaP{2}$-complete, then the checking part $C$ must contain disjunctive
rules unless $C$ has feedback to the guessing part $G$.

\nop{
In particular, if the above program is rewritten to eliminate such
feedback then, like in the program $\p{1}$ above, further disjunctive
rules must be added. Further examples of programs in which the
checking part necessarily contains disjunctive rules can be found in
\cite{eite-etal-97f}.
}

Finally, note that STRATCOMP can not be expressed by a fixed normal logic
program uniformly on all collections of facts {\tt
produced\_by$(p,c1,c2)$} and {\tt controlled\_by}$(c,c1,$ $c2,c3)$
(unless $\NP=\SigmaP{2}$, an unlikely event).

\subsection{Specifics}
                                        

\dlv is the result of putting theoretical results into
practice. It is the first system supporting answer set semantics for
full disjunctive logic programs with negation, integrity constraints,
queries, and arithmetic built-ins.

The semantics of the language is both precise and
intuitive, which provides a clean, declarative, and easy-to use
framework for knowledge representation and reasoning.

The availability of a system supporting such an expressive language in
an efficient way is stimulating AI and database people to use
logic-based systems for the development of their applications.

Furthermore, it is possible to formulate translations from many other
formalisms to \dlv's core language, such that the answer sets of the
translated programs correspond to the solutions in the other
formalism. \dlv incorporates some of these translations as
frontends. Currently frontends for diagnostic reasoning, SQL3,
planning with action languages, and logic programming with inheritance
exist.

We believe that \dlv can be used in this way -- as a core engine --
for many problem domains. The advantage of this approach is that
people with different background do not have to be aware of
\dlv's syntax and semantics.

\subsection{Users and Usability} 

Prospective users of the \dlv core system should have a basic
knowledge of logics for knowledge representation. As explained in the
previous section, if a frontend for a particular language exists, a
user need not even know about logics, but of course knowledge about
the frontend language is still required.

Currently, the \dlv system is used for educational purposes in
courses on Databases and on AI, both in European and American
universities. It is also used by several researchers for knowledge
representation, for verifying theoretical work, and for performance
comparisons.

Furthermore, \dlv is currently under evaluation at CERN, the European
Laboratory for Particle Physics located near Geneva in Switzerland and
France, for an advanced deductive database application that involves
complex knowledge manipulations on large-sized databases.

\section{Evaluating the System}


\subsection{Benchmarks}

It is a well-known in the area of benchmarking that the only really
useful benchmark is the one where a (prospective) user of a system
tests that system with exactly the kind of application he is going to
use.

Nevertheless, artificial benchmarks do have some merits in developing
and improving the performance of systems. Moreover, they are also very
useful in evaluating the progress of various implementations, so there
has been some work in that area, too, and it seems that \dlv compares
favorably to similar systems \cite{eite-etal-98a,janh-etal-2000}.


Also for the development of some deductive database applications \dlv can
compete with database systems. Indeed, \dlv is being considered by CERN
for such an application which could not be handled by other systems.

\subsection{Problem Size} 

As far as data structures are concerned, \dlv does not have any real
limit on the problem size it can handle. For example, we have verified
current versions on programs with 1 million literals in 1 million rules.

Another crucial factor for hard input are suitable heuristics. Here we
already have developed an interesting approach \cite{fabe-etal-99b} and
are actively working on various new approaches.

To give an idea of the sizes of the problems that \dlv can currently handle,
and of the problems solvable by \dlv in the near future,
below we provide the execution times of
a number of hard benchmark instances reporting also the improvements
over the last year.

\medskip

\noindent{\small
\begin{tabular}{@{\extracolsep{\fill}}lrrrrrr}
\hline
\it Problem & \it Jul.\ '98 & \it Feb.\ '99 \nop{& \it Apr.\ '99} & \it Jun.\ '99 & \it Nov.\ '99\\
\hline
3COL$^a$                 &     $>$ 1000s &         26.4s \nop{&         13.7s} &          2.1s &          0.5s\\
HPATH$^b$     &     $>$ 1000s &     $>$ 1000s \nop{&        627.0s} &         10.8s &          0.3s\\
PRIME$^c$     &      --- \nop{$^g$} &         21.2s \nop{&         13.1s} &         10.2s &          0.8s\\
STRATCOMP$^d$  &         54.6s &          8.0s \nop{&          7.4s} &          6.9s &          5.4s\\
BW P4$^e$       &     $>$ 1000s &     $>$ 1000s \nop{&     $>$ 1000s} &         32.4s &          6.3s\\
BW Split P4$^f$ &     $>$ 1000s &     $>$ 1000s \nop{&        279.4s} &         10.5s &          2.3s\\
\hline
\end{tabular}

\medskip 

{\footnotesize
$^a$find one coloring of a random graph \\
\hspace*{\parindent}\ with 150 nodes and 350 edges

$^b$find one Hamiltonian Path in a random graph \\
\hspace*{\parindent}\ with 25 nodes and 120 arcs

$^c$find all prime implicants of a random 3CNF \\
\hspace*{\parindent}\ with 546 clauses and 127 variables

$^d$find all strategic sets a randomly chosen company\\
\hspace*{\parindent}\ occurs in (71 companies and 213 products)

$^e$find one plan of length 9 involving 11 blocks

$^f$linear encoding for $e$
}


\newcommand{\SortNoOp}[1]{}

\end{document}